\pdfoutput=1
\documentclass[runningheads]{llncs}
\usepackage{hyperref}
\usepackage{comment}
\usepackage[T1]{fontenc}
\usepackage{graphicx}
\usepackage{subfig}
\usepackage{multirow}
\usepackage{array}
\usepackage{color}
\usepackage[numbers]{natbib}
\usepackage{graphicx}
\usepackage{algpseudocode}
\usepackage{algorithm}
\usepackage{todonotes}
\usepackage{booktabs} 

\begin{document}
%
\title{Fine-tuning of explainable CNNs for skin lesion classification based on dermatologists' feedback towards increasing trust}

\titlerunning{Fine-tuning of explainable CNNs for skin lesion classification}
%
\author{Md Abdul Kadir\inst{1}\orcidID{0000-0002-8420-2536} \and
Fabrizio Nunnari \inst{1}\orcidID{0000-0002-1596-4043} \and
Daniel Sonntag\inst{1,2}\orcidID{0000-0002-8857-8709}}
 \authorrunning{MA Kadir et al.}

\institute{
German Research Center for Artificial Intelligence (DFKI), Saarland Informatics Campus, D3 2, 66113 Saarbrücken, Germany \and
University of Oldenburg, Marie-Curie Str. 1, 26129 Oldenburg, Germany
\email{\{info,abka03\}@dfki.de}\\
\url{http://www.dfki.de}
}
\maketitle              
\begin{abstract}

In this paper, we propose a CNN fine-tuning method which enables users to give simultaneous feedback on two outputs: the classification itself and the visual explanation for the classification.
We present the effect of this feedback strategy in a skin lesion classification task and measure how CNNs react to the two types of user feedback.
To implement this approach, we propose a novel CNN architecture that integrates the Grad-CAM technique for explaining the model's decision in the training loop.
Using simulated user feedback, we found that fine-tuning our model on both classification and explanation improves visual explanation while preserving classification accuracy, thus potentially increasing the trust of users in using CNN-based skin lesion classifiers.
\keywords{Skin lesion \and XAI (Explainable Artificial Intelligence) \and SENN (Self Explainable NN) \and XIL (Xplanatory Interactive Learning)} 
\end{abstract}

\section{Introduction}
A Convolutional Neural Network (CNN) can detect malignant skin lesions \cite{JAIN2015735};
﻿however, it cannot produce by default the explanation behind a prediction. In image classification,
﻿there are several ways to explain a prediction \cite{selvaraju2016grad-cam,zhouetal2016learning,petsiuk2018rise}. Nonetheless, in some cases, the explanations can
﻿be misleading, and by default the network does not provide the flexibility of learning from a given correction.
As a result, the acceptance of such algorithms in the medical domain is quite rare. 
Differently, in the same situation, given the availability of correct feedback, decision-makers can perceive the reason for the mistake and take the necessary action to avoid the same kind of mistake in the future. These limitations of neural networks fall under the category of lack of interactivity. 

﻿
﻿
﻿
﻿
In the domain of image classification, researchers introduced visual
﻿techniques to introduce explainability in deep learning-based image classification. They
﻿visualize the discriminatory regions of an image based on specific class identity \cite{selvaraju2016grad-cam}.
﻿Highlighting class discriminative regions in an image is an example of explainability.
﻿
This helps in spotting biased CNNs that wrongly identify the location of interest used to achieve the classification result.
For example, sometimes classification performance is good, but the model classifies based on unrelated features.
How can we tell the CNN that it is looking at the wrong image region and at the same time improve its classification accuracy?


﻿
﻿

\begin{figure}[t]
    \centering
    \includegraphics[width=0.8\textwidth]{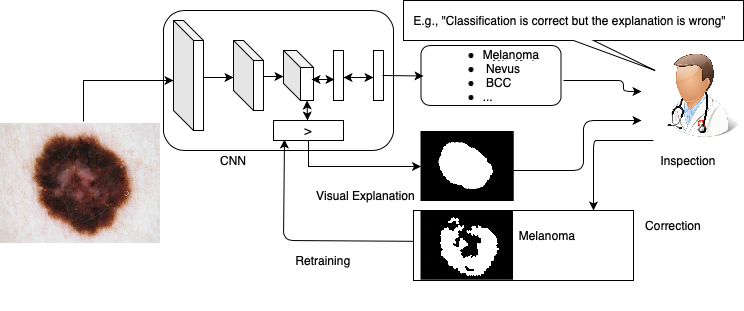}
    \caption{Method overview: a pre-trained classification model produces classification result and visual explanation. Our interactive method allows a specialist to provide, if needed, corrections on both of them. Corrected information is then used to fine-tune the model on both output types.}
    \label{fig:method_overview_data}
\end{figure}

We address this problem through the user-feedback approach is depicted in Fig. \ref{fig:method_overview_data}.
We assume to start from a model that has been trained solely on classification data.
Whenever a new sample is passed for classification, the resulting classification and visual explanation can be, independently, correct or wrong. If the correction is applied to the classification results, the solution is rather trivial. One simply needs to collect a batch of corrected labels and use them to perform transfer learning, or fine-tuning, on the starting model. Technically, this is done with the same data formats used for the original training.
However, the challenge comes when the correction is applied to the visual explanation image.

In this paper, we propose a novel fine-tuning method that accepts human-corrected visual explanation images as part of the forward/backward propagation loop.
On the technical level, our contribution consists of: i) the implementation a modified differentiable version of the Grad-CAM \cite{selvaraju2016grad-cam} explanation technique (so that corrected explanation images can be included in the back-propagation phase), and ii) the definition of a training procedure that takes into account simultaneously the classification and the visual feedback.
With this approach, the difference between the original explanation image and the corrected version can be used as additional term of a loss function that includes, together with a classification error, also a visual explanation discrepancy.

Experiments are conducted in the domain of skin lesion classification, simulating how dermatologists could possibly identify wrong classifications and explanations and provide correction feedback.

\section{Related work}
﻿This section presents the researches related to explainability and
﻿interactivity in the machine learning and the deep learning domains.
﻿At the beginning phase of their development, machine learning and
﻿deep learning algorithms were black-box models, except for linear
﻿and tree-based models. Black-box models only predict, but they do
﻿not present the reason behind their prediction.
﻿
There are two ways for explanation: explaining
﻿the model itself, or explaining the reason for a specific prediction.
﻿They are known as global and local methods. In this research, we will
﻿only focus on the local methods because the research aims to improve
﻿the explanation of a sample-based prediction.
﻿First, we will cover the different explanation generation techniques
﻿of CNN-based and typical ML-based image classifiers. After that,
﻿we will discuss existing interactive methods where users provide
﻿feedback to NNs/AI algorithms to improve accuracy. We will also
﻿present the limitation of these methods.

\subsection{Class activation mapping}
A class activation technique generally works in CNNs. In this technique,
﻿the discriminatory classification features are extracted from the
﻿activation of any convolutional layer.
﻿Generating class activation map (CAM) is a technique for localizing
﻿class-specific significant features used in explaining convolutional
﻿neural networks. CAM has a remarkable localization capability.
﻿Zhou \cite{zhouetal2016learning} describe a procedure for generating CAM
﻿using global average pooling (GAP) on convolution layers. GAP
﻿is the weighted sum of the convolutional feature map. In some
﻿CNN, a neural decision is made from the weighted sum of the
﻿GAP outputs. According to the authors, we can spot important image regions by linearly combining the weights of the output layer with the activation of the last convolution layer.
Selvaraju \cite{selvaraju2016grad-cam} propose an explanation technique known
﻿as Grad-CAM, which is an extended version of CAM. It utilizes any target
﻿class flow gradients through the final convolution layer.
﻿Grad-CAM++ \cite{chattopadhayetal2018} is an extended version of Grad-CAM.
﻿In "Improved Visual Explanations for Deep Convolutional Networks",
﻿Chattopadhay  \cite{chattopadhayetal2018} propose this generalized method.
﻿It can produce improved visualization behavior of CNNs' predictions and performs better in visualizing multiple instances of an object
﻿during a classification. It produces an explanation similar to
﻿how Grad-CAM produces an explanation, but the only difference
﻿is that it only considers the positive gradient of the output class.
﻿
Barata \cite{Barata_2019_CVPR_Workshops} 
﻿proposed a hierarchical CNN-LSTM  attention model that uses
﻿hierarchical information about classes and then produces
﻿attention mapping and hierarchical classification results.
﻿The attention map hints at how the classification
﻿algorithm looks at the objects in an image.


In this research we leverage class activation mapping to generate post-hoc explanation of our model while tuning. Any gradient based explanation technique can adapted to our method.

﻿\subsection{Interactive and explainable AI }
﻿Teso and Kersting \cite{EAL} argue that interactive learning places
﻿the user into the loop, but the learner stays as a black-box for the user.
They also suggest a novel explanatory interactive learning (XIL) framework that can overcome the limitation of interactive learning. Moreover, it can help the user gain trust in the 
﻿learner by introducing completeness, directability, and understandability. 
﻿In XIL, a user gives feedback to the learner's output in
﻿an active learning manner when required. The proposed framework
﻿utilize LIME \cite{10.5555/1248547.1248608} as a local explainer and
﻿an additional component. They call the framework Caipirinhas (CAIPI).
﻿They use three functionalities, labeling the unlabeled data using user input,
﻿fitting the model on labeled and unlabeled features, and explaining
﻿a prediction using the local explainer. By introducing counter example, authors allow CAIPI to learn
﻿from the user's feedback on the label and the explanation. 
Counterexamples are nothing but original
﻿input images with randomized irrelevant regions. There are three scenarios during
﻿the interaction between the learner and the user: the prediction and the
﻿explanation are correct, both are wrong, or only the label is correct.
﻿CAIPI focuses mainly on the last. It trains itself from the user's feedback.
﻿After the evaluation, the authors see that the trust/distrust of the
﻿user increase based on the interactions.
﻿CAIPI's performance increases due to the feedback explanation.
﻿However, the counterexample requires more disk space to store
﻿and GPU performance to retrain the model.
﻿The main difference between CAIPI and this research is how we train the network on feedback.
﻿CAIPI augments the original training data based on user feedback and then fine-tune or
﻿train a new model. There is no change in the objective function of the model.
﻿Also, storing original training data is necessary for retraining. However,
﻿the approach we follow in this research only requires post-deployment test images
﻿and feedback. Initial training data is unnecessary. 

﻿

﻿According to Teso \cite{ISENN}, the Explanatory Active Learning (EAL)
﻿\cite{EAL} algorithm depends on a post-hoc explainer, and it can generate
﻿a fragile and unfaithful explanation. He says the self-explainable active
﻿learning model is a solution to that. It is a combination of active
﻿learning and self-explainable neural networks (SENNs) \cite{SENN}. Ghai \cite{XAL} introduce Explainable Active Learning (XAL)
﻿in An Empirical Study of How Local Explanations Impacts Annotator
﻿Experience in 2020.
﻿Stuntebeck \cite{healthsense} propose a human-in-loop machine learning framework.
﻿This framework collects data from the patient using health sensors and
﻿trains a machine learning model on that data.  Sometimes, due to the inefficiency
﻿of the sensors, the model prediction becomes wrong.  To overcome this problem,
﻿they involve the user in the learning loop.  Occasionally, the model gives the 
﻿prediction, and the user gives feedback on the prediction by comparing what
﻿they are experiencing.  Based on the feedback, the model tune itself.
﻿This framework is similar to this research involving humans in the learning loop.
﻿However, the feedback in this framework is only a yes or no decision.
﻿Holzinger \textit{et al.}\ \cite{glassbox} argues that, while automatic ML suffers in performance
﻿because of insufficient training data, it is also true that interactive
﻿ML has the flexibility to allow a user to select suitable features heuristically
﻿from a vast search space.  As a result, it can reduce the complexity of NP-hard 
﻿using outside knowledge (Human intervention).  The authors demonstrated the effectiveness
﻿of interactive machine learning and showed how to open a black-box technique
﻿to a glass-box one, enabling humans to interact with an algorithm.
﻿In the Skincare project, Sonntag \textit{et al.}\ \cite{sonntag_skincare_2020} describe
﻿the functionality and interface of an interactive decision support system for
﻿differential diagnosis of malignant skin lesions. The methods in the report give generic ideas and importance for interactive machine learning.

Besides lesion segmentation \cite{jafari_et_al_2016}, several pieces of research are published in skin lesion
﻿classification \cite{lopez2017skin}.
﻿We see that there are several approaches for skin cancer classification.
﻿Many of the methods have human-level accuracy. However, the application of these methods
﻿is constantly challenged by critique for legitimate reasons; for example,  the proposed
﻿classifiers are black-box models, need training updates on new data, and lack
﻿interactivity with humans and the environment. So, in this research, we explore the interactive side of skin cancer classification.

\section{Methods}

Our method consists of a deep neural architecture that explains decisions to users and gives them the possibility to perform corrections that improve the model's performance.
Fig. \ref{fig:method_overview_data} provides an overview of the system.

From the left side, we see a skin lesion fed into a convolution neural network, which predicts the class of the lesion and shows the areas that mostly contributed to the result to a dermatologist (on the right side).
If the dermatologist is not convinced of the classification result or with its explanation, s/he gives a feedback. Based on the feedback, we re-train the model.

The remaining of this section explains the details of the implementation and describes how we simulated the feedback of dermatologists to validate our approach.

\subsection{Integrating the explanation with the classification results}

Out method starts from a pre-trained CNN which able to perform image classification.
A classifier based on convolutional networks is typically composed of two functional blocks: the convolutional layers (CL) block and the the fully connected layers (FCL) block. Images ($X$) are fed into the CL at first, and the outputs of CL are flattened and passed through the FCL to generate the output ($y$).
The output of the last convolutional layer is analyzed by explanation local techniques like Grad-CAM to produce a saliency map.
A saliency map is a grey-scale image, with the same resolution of the convolutional layer, whose pixels with higher luminance are associated to the areas of the image that mostly contributed to the classification, while areas with dark pixels were mostly ignored.

The main idea of the method proposed in this paper, is to provide the results of a classification in terms both of classification result \emph{and} saliency map to an expert, ask him to perform corrections, and use the corrected result (or a batch of corrections) for further fine tuning of the model. The fine tuning will take into account both the class \emph{and} the saliency map into a new composite loss function. This is done by attaching to the original model a differentiable branch able to extract the saliency map during a forward pass.

However, from a human perspective, correcting a grey-scale image might be too difficult and time consuming. In fact, in the context of skin cancer detection, many datasets available for the masking tasks provide \emph{binary} masking images.
Hence, we accommodate users' usability by converting greyscale saliency maps into binary \emph{explanation maps}, where the users corrections consists only in switching pixels status between black and white, or viceversa.
The conversion from saliency into explanation map ($y_{exp}$) can be performed by simple thresholding~\cite{nunnari_on_the_overlap}. 

The new loss function of the extended model has two components, classification loss and explanation loss. The classification loss is the loss of a pre-trained network. However, the explanation loss is a newly introduced function, and it punishes the overall cost based on the difference between generated explanation and user feedback on explanation.
The loss function can be written as 

\begin{equation}
L = (1-\lambda)L_{cls}(y, \hat{y}) + \lambda L_{exp} (y_{exp}, \hat{y}_{exp}) 
\end{equation}

Here $L_{cls}$ and $L_{exp}$ are the classification and explanation loss respectively. The hyper-parameter $\lambda$ modulates the contribution of the two terms, and can be set during the model tuning. The two losses are defined as:

\begin{equation}
L_{cls}(y, \hat{y}) = -\sum_i \hat{y_i} \log (y_i)
\end{equation}

\begin{equation}
L_{exp}(y_{exp}, \hat{y}_{exp}) = \sum_i J(y_{exp_i}, \hat{y}_{exp_i})
\end{equation}

Here, $L_{cls}$ is the well-know categorical cross-entropy, while $L_{exp}$ is the result of the Jaccard index between the output explanation map and the corrected one. An explanation map $y_{exp}$:

\begin{equation}
y_{exp} = T(A_y)
\end{equation}

is the thresholded version of the result of the Grad-CAM saliency map $A_y$, which is defined as:

\begin{equation}
A_y = ReLU (\sum a_k^c A^k)
\end{equation}

where $A^k$ is the activation in a convolution layer $k$. Given $Z$ as the total number of pixel in $A^k$ (with resolution $i \times j$), for a given class $c$,
$a_k^c$ is computed as:

\begin{equation}
a_k^c = \frac{1}{Z} \sum_i \sum_j \frac{\Delta y^c} {\Delta A_{ij}^k} 
\label{eq:ak}
\end{equation}

\subsection{Implementation details}

Algorithm \ref{alg:imp} reports a formalized description of the approach. The main aspect that is worth an explanation, is that the Grad-CAM algorithm needs an alteration of the last layer in order to compute its saliency map. For this reason, each iteration of the fine-tuning algorithm requires two forward passes. The first, for the classification, is performed on the original model, while the second is performed on a temporary copy of the model, which is modified to get the saliency map and destroyed and the end of the batch iteration.

\begin{algorithm}
\caption{Training the self-explainable model}\label{alg:imp}
\begin{algorithmic}
\scriptsize
\Require $e$ \Comment{e: number of epoch}
\Require $m = F(\theta)$ \Comment{$m$: pre-trained model }\
\Require $X, Y, Z$ \Comment{$X$: input sample, $Y$: label  , $Z$: binary mask (i.e. explanation ground truth)}
\Require $N$ \Comment{$N$: total samples}
\Require $L_{cls}$ \Comment{$L_{cls}$: classification loss function}
\Require $L_{exp}$ \Comment{$L_{exp}$: explanation loss function}
\Require $\gamma$ \Comment{$\gamma$: learning rate}
\State $i=1$
\While{$ i \leq e $}\
\State $n = 0$
\While{$n \leq N$}\
\State $ m_g \gets deepcopy(m)$\
\State $a_k^c \gets \frac{1}{R} \sum_p \sum_q \frac{\delta m_g(x)[c]} {\delta A_{ij}^k}$\ \Comment{$\frac{\delta m_g(x)[c]} {\delta A_{ij}^k}$ : class-specific gradient on layer $k$ for image $x$}
\State $A^k, \hat{y} \gets m (x)$ \Comment{$A^k$: receptive field of layer $k$, $\hat{y}$: predicted class} 
\State $S \gets ReLU(A^k \times a^c_k)$ \Comment{$S$: saliency map}
\State $\hat{y_{exp}} \gets th(S, t)$ \Comment{t: threshold value}\
\State $L(\theta) \gets (1-\lambda) L_{cls}(y,\hat{y} ) + \lambda L_{exp}(z, \hat{y_{exp}})$ \Comment{$\lambda$: hyper-parameter for loss balance}
\State $ \theta \gets \theta - \gamma \nabla L(\theta) $ \Comment{$\nabla L(\theta)$: gradient with respect to loss}
\EndWhile
\EndWhile
\end{algorithmic}
\end{algorithm}

\subsection{Simulation of user feedback}

\begin{figure}
    \centering
    \includegraphics[width=0.7\textwidth]{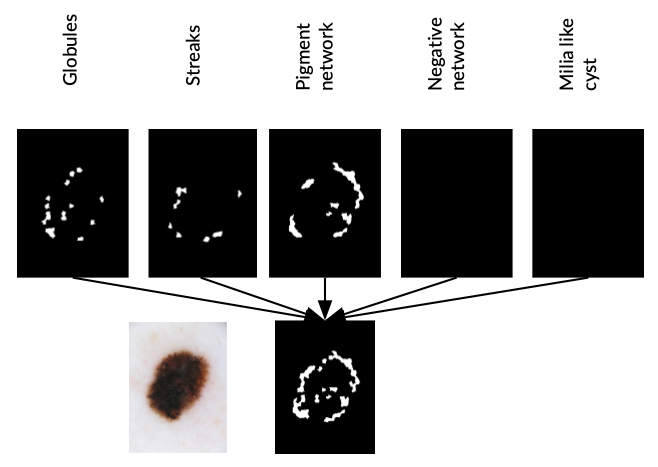}
    \caption{The union of the feature maps. }
    \label{fig:union}
\end{figure}

Getting user feedback for an experiment is very costly and time-consuming especially in the medical domain.
Hence, instead of performing tests with real users, we exploited an existing dataset of skin lesion images associated to both classes and explanation maps, and used it to simulate users' feedback.
The ISIC 2018 \cite{skincaredata} dataset of skin lesion attributes contains images of skin lesions and masks of five different attributes, pigment network, negative network, streaks, milia like cyst, and globules (Fig. \ref{fig:union}). These attribute maps are binary masks locating the different attributes. The union of all of these attribute maps provides a comprehensive indication of what are the pixel areas that would lead a human practitioner towards his/her decision.

\begin{figure}[t]
\centering
\begin{tabular}{cccc}{
﻿\subfloat[NV]{\includegraphics[width =1in, height=1in]{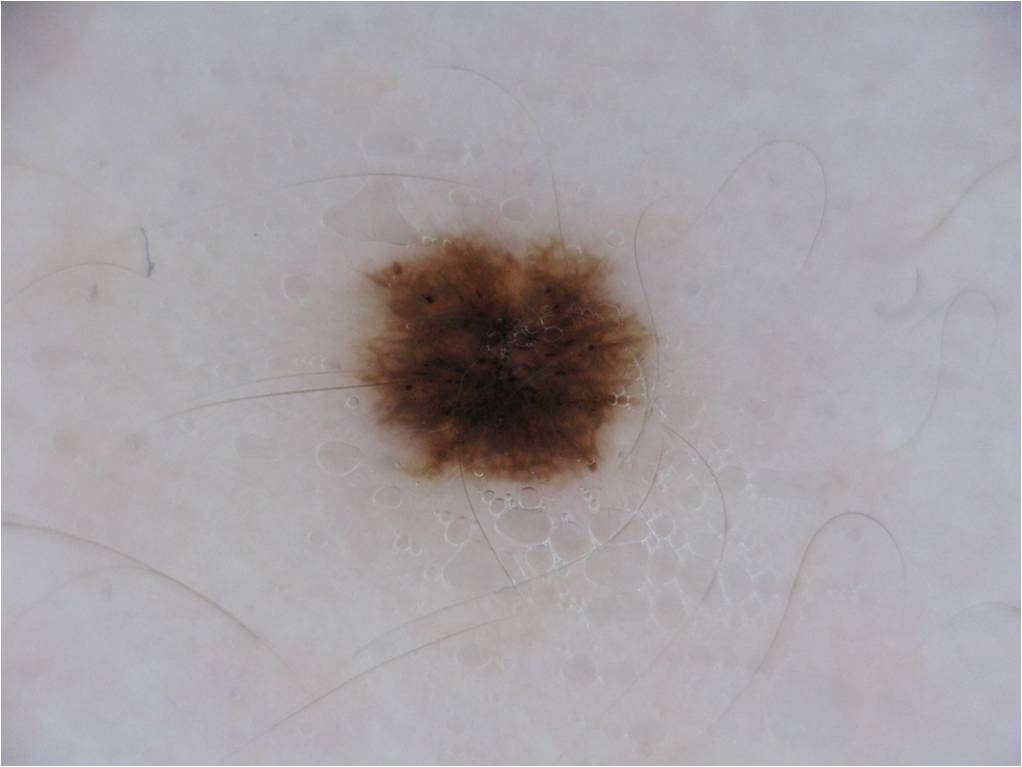}}} &
\subfloat[MEL]{\includegraphics[width = 1in, height=1in]{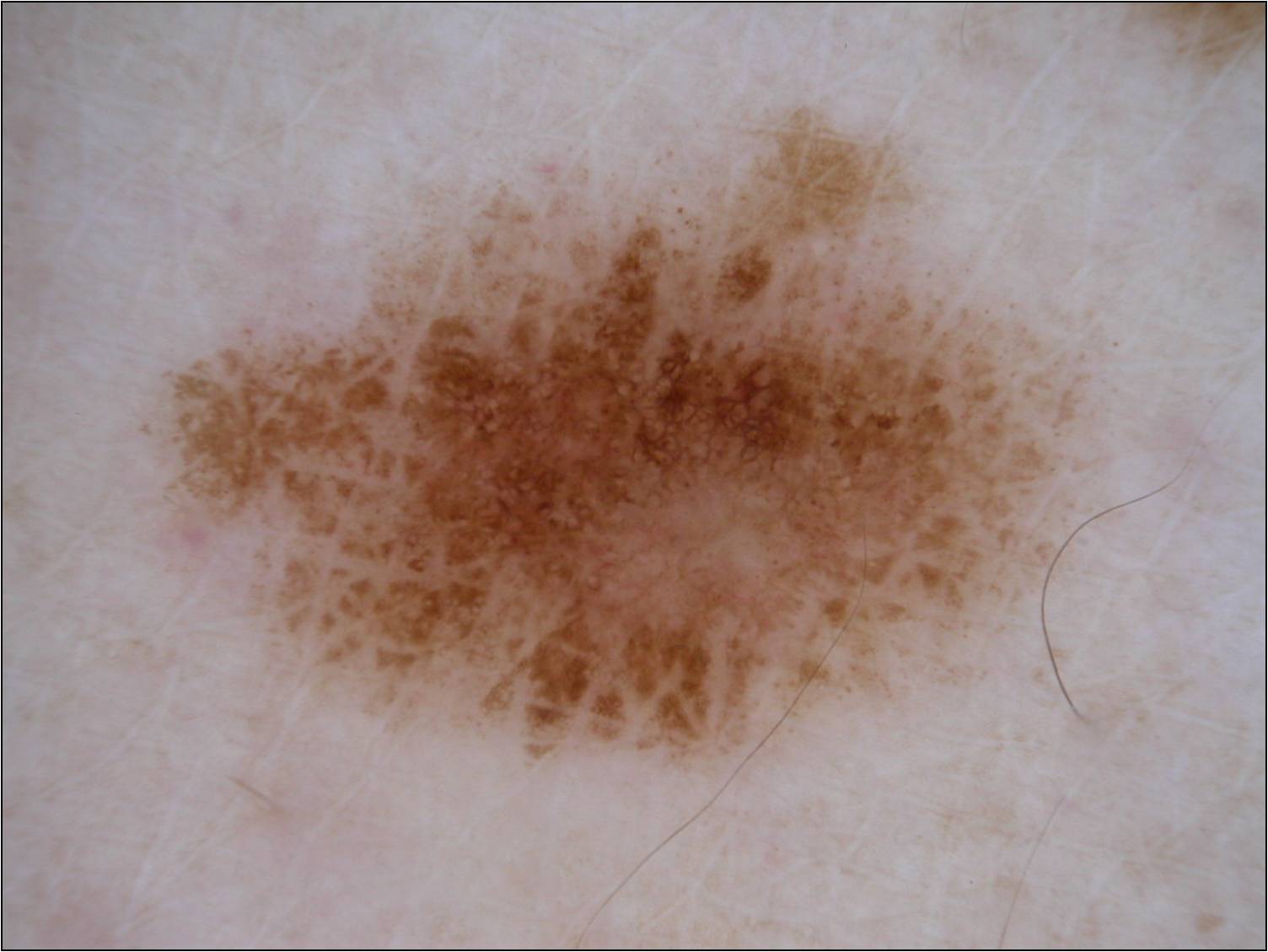}} &
\subfloat[BKL]{\includegraphics[width =1in, height=1in]{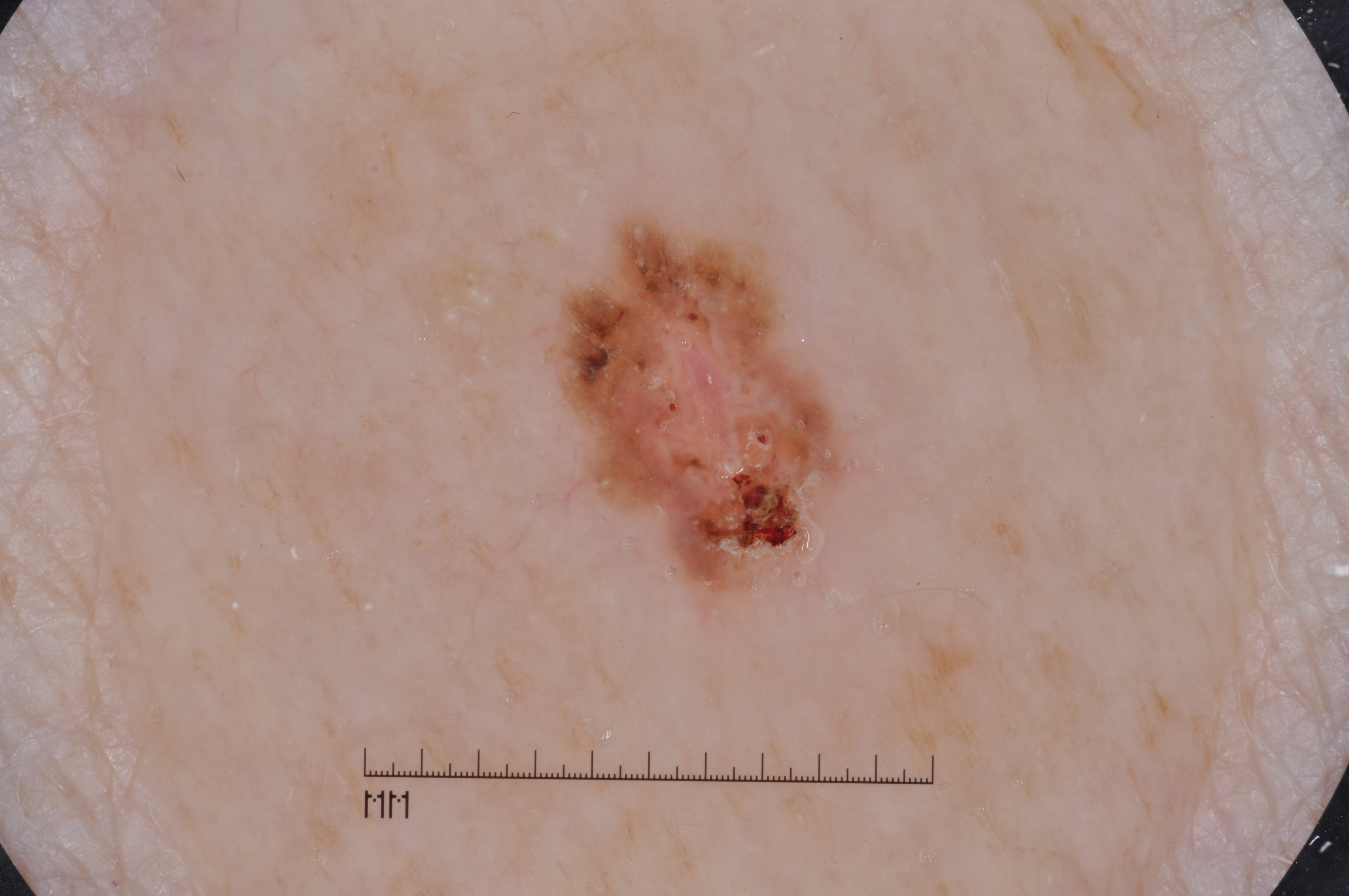}}\\
\end{tabular}
\caption{Sample images from ISIC2018 task 2 dataset}
\label{fig:sample_image}
\end{figure}

The ISIC2018 dataset (Fig. \ref{fig:sample_image}) is imbalanced. Preliminary experiments showed us that such imbalance affects the results of the simulation. Hence, we equalize the data per class by upsampling in the simulation set. We know that the maximum sample belongs to the nevus class (1951 samples), but there are 437 examples for the MEL class and only 172 samples for the BKL class. We increase the number of the MEL and BKL samples to 1951 samples by coping them randomly.

\section{Experiments and results}

This sections described the two experiments conducted with our approach: the first aiming at understanding the contribution of the explanation loss function on a batch of data feedback, and the second aiming to analysing the behavior of the model when data feedback are provided in smaller batches.

﻿\subsection{Training on different loss functions}
﻿We use three different loss functions in the experiment: classification loss, explanation loss,
﻿and combined loss. The purpose of training the model on classification loss is to see how models
﻿perform when feedback is given only to classification data. During the training on classification, the model gets 
﻿feedback only on classification output. So, the loss function only contains the classification
﻿part. We observe training and validation loss and average Jaccard index during the tuning process.
﻿Differently, the objective of using explanation loss is to see how the model performs when
﻿there is only explanation feedback. The explanation loss is the Jaccard loss which compares the
﻿predicted and feedback explanations. Finally, the combined loss combines both losses by a
﻿regularization parameter $\lambda$, which balances the two losses. The purpose is to see how
﻿classification and explanation feedback improve model performance.

We keep 10\% of the data untouched to check the model performance and to see if the model learns from feedback, another 10\% for validation and the remaining 80\% for fine tuning the model through user simulation.


Table \ref{tab:main_table} presents the final result of the simulation.
The baseline model performs 0.71 average sensitivity on the original test set (from ISIC 2019), and 0.73 when tested on the full simulation set (ISIC 2018).
When performing a test on the 10\% of the simulation set, using only the classification loss leads to 0.70 sensitivity and a Jaccard index as low as 0.10.
Combining the classification and the explanation loss keeps the 0.70 sensitivity and increases the Jaccard index to 0.127, hence increasing the explanation power of the model. When testing using only the explanation loss, the Jaccard index boosts to 0.18, but the sensitivity drops to 0.66.

Hence, it seems that including an explanation loss term doesn't affect the classification capabilities of the mdoel but it is able to improve its explanation power.

\begin{table}[t]
    \scriptsize
    \begin{center}
        \begin{tabular}{ p{1.1cm}|p{2.1cm}|p{.62cm}|p{1.7cm}|p{.6cm}|p{.6cm}|p{.7cm}|p{.7cm}|p{.7cm}|p{0.8cm}|p{1cm}}
        \toprule
        Model &Train data &Loss & Test set &$\lambda$&Test acc& \multicolumn{3}{c|}{Sensitivity} & Avg sensitivity & Avg J-mean\\
        \midrule
        &&&&&&MEL&NV&BKL&&\\
        Baseline & ISIC 2019 &$L_{cls}$ & ISIC2019 2.5K test & n/a &	0.71&0.61&0.74&0.65&0.71&n/a\\
        Baseline &ISIC 2019&	$L_{cls}$& 100\% Sim Set& n/a&0.73&0.53&0.77&0.88&0.73 & 0.10\\  
        \midrule
        Sim-model& 80\% of Sim Set &$L_{cls}$&10\% Sim Set&0&0.74&0.74&	0.75&0.61&0.70 & 0.106 (0.08)\\
        Sim-model&  80\% of Sim Set&$L_{exp}$&10\% Sim Set&1&	0.76 & 0.49 & 0.85 & 0.64 & 0.66 & \colorbox{green}{0.18} \colorbox{green}{(0.13)}\\
        Sim-model& 80\% of Sim Set&$L_{cls}$ \& $L_{exp}$&10\% Sim Set& 0.3&0.64&	0.78&0.59&0.72&	0.70	& 0.127 (0.10)\\
        \bottomrule
        \end{tabular}
        \caption{Complete result of simulation on upsampled samples. We see that there is improvement of average Jaccard index when the explanation loss is included in the loss computation.}
        \label{tab:main_table}
    \end{center}
\end{table}

\subsection{Sliced simulation}
\begin{figure}[t]
    \centering
    \subfloat[\centering Accuracy vs slices ]{{\includegraphics[width=0.49\textwidth]{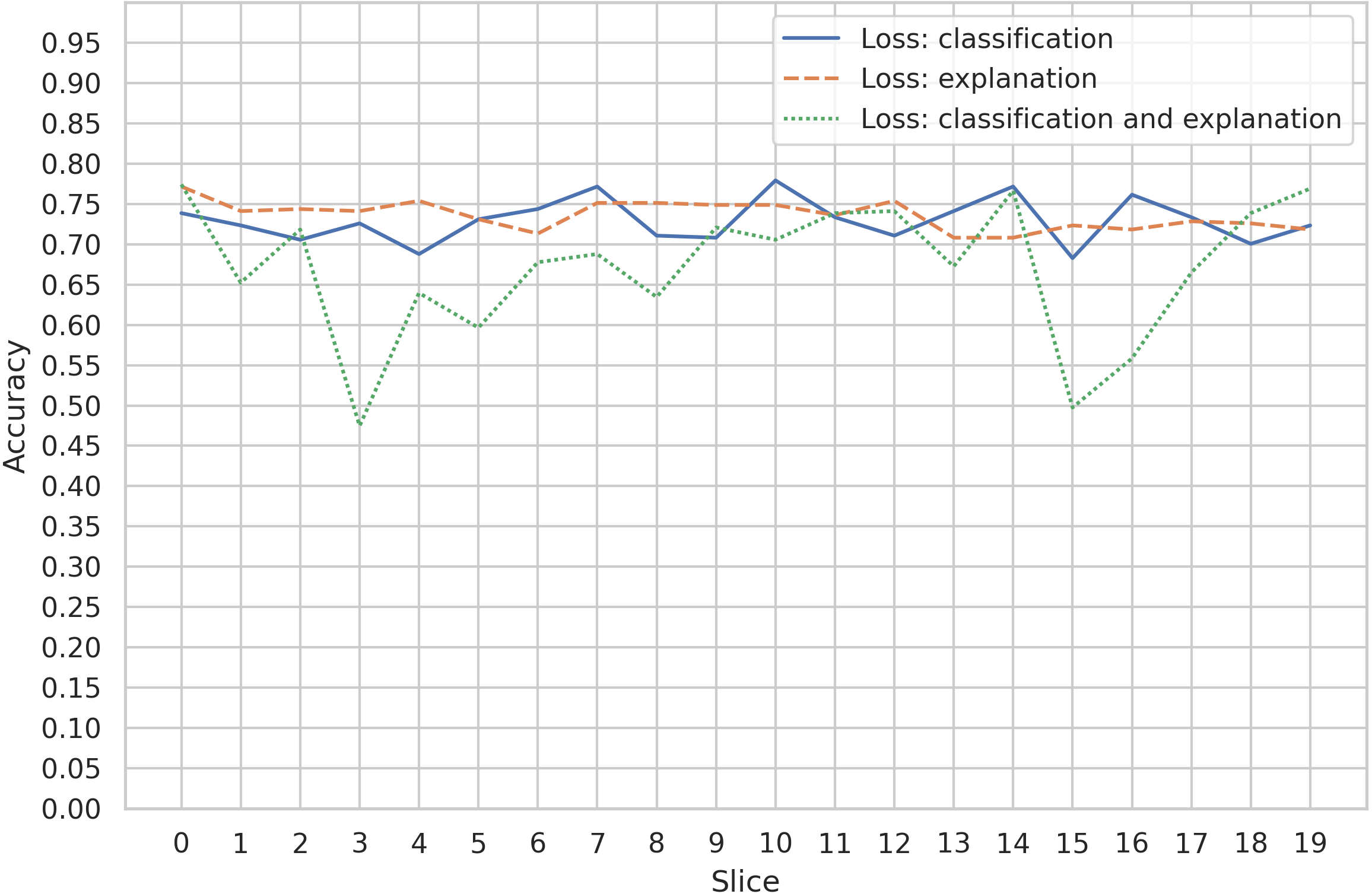} }}%
    \subfloat[\centering Jaccard score vs slices]{{\includegraphics[width=0.49\textwidth]{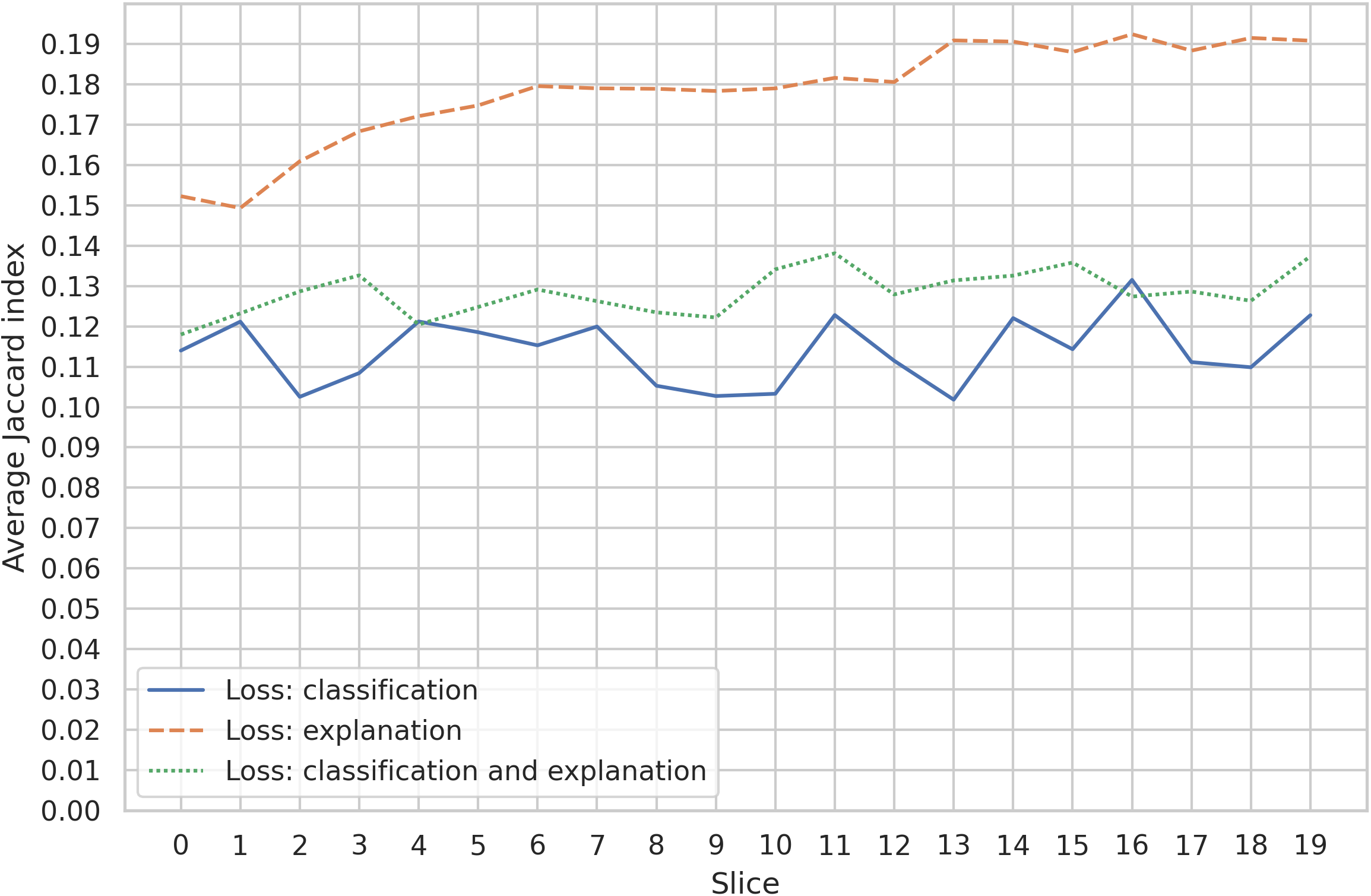} }}%
    \caption{An overview of performance in each epoch}%
    \label{fig:slice_exp}%
\end{figure}

In this section, 
﻿
﻿we present how a gradual provision of tuning data improves our explainable model's performance.
﻿Fig. \ref{fig:slice_exp}(a) shows how model accuracy increases over an increasing amount of data. The x-axis represents data slices. Each slice has 238 samples. We have a total of 20 slices, and the model is iteratively fine-tuned on all the slices.
We plotted the accuracy evolution for the three kinds of loss functions: classification only, explanation only, and combined.
For the first two cases, we see that the accuracy has no significant improvement over the slices, while we see more accuracy fluctuation when combining classification and explanation loss functions. However, overall, accuracy remains stable at the end of the slices provision.

Fig. \ref{fig:slice_exp}(b) presents the change in the average Jaccard index over the slices. Looking at the classification loss function's graph, we see that the average Jaccard index is not increasing.
There is a slight improvement when we use combined loss. However, we see notable improvement of the average Jaccard index while using the explanation loss function.
Table \ref{tab:sliced_table} shows the summary of the sliced simulation. After tuning the model on the first slice, we see that the average Jaccard index is 0.15.
However, after the 20th slice, we see that the Jaccard index increases to 0.19. On the other hand, the accuracy reduces to 0.72 from 0.77 by keeping the average sensitivity the same when using the explanation loss function. 
These results conclude that model explanation performance increases without reducing classification accuracy when the model gets correction feedback in smaller chunks.

From those experiments, it seems that there is a slight better ending performance in the model after training over 20 smaller feedback slices rather than on the full 80\% of the simulation set. However, the margins are small and this observed behaviour needs to be further analysed on bigger datasets to be confirmed.

\begin{table}[b]
    \scriptsize
    \begin{center}
       \begin{tabular}{p{3.2cm}|p{1cm}|p{2.1cm}|p{1cm}|p{2cm}|p{2cm}}
            \toprule
            \textbf{Loss function} & \textbf{Slice \#} &  \textbf{Test set} & \textbf{Test acc} & \textbf{Test avg. sensitivity} & \textbf{Avg. Jaccad index(sd)}\\
            \midrule
            $L_{cls}$ & 0 & 20\% of simSet & 0.74 & 0.70 & 0.11 (0.09)\\
            $L_{cls}$ & 19 & 20\% of simSet & 0.72 & 0.69 & 0.12 (0.10)\\
            $L_{exp}$ & 0 & 20\% of simSet & 0.77 & 0.66 & 0.15 (0.12)\\
            $L_{exp}$ & 19 & 20\% of simSet & 0.72 &0.67 & \colorbox{green}{0.19 (0.15)}\\
            $0.70 \times L_{cls} + 0.3 \times L_{exp}$ & 0 & 20\% of simSet & 0.77 & 0.66 & 0.12 (0.10)\\
            $0.70 \times L_{cls} + 0.3 \times L_{exp}$ & 19 & 20\% of simSet & 0.76 & 0.70& 0.14 (0.11)\\
            \bottomrule
        \end{tabular}
       \caption{Complete result of sliced simulation. As model gets more explanation and classification feedback, there is an improvement of average Jaccard index as well as of classification accuracy (last line).}
       \label{tab:sliced_table}
     \end{center}
\end{table}


%
%


%
%
%
\section{Conclusion}

We presented an approach for increasing the performances of CNN-based skin cancer classification by including users feedback in a post-train, fine-tuning stage.
This approach transforms a black-box VGG16 model into a self-explainable neural network (SENN), which classifies (categories) and explains (saliency maps) at the same time.
The implementation consists of augmenting a pre-trained VGG16 architecture with a differentiable implementation of the Grad-CAM algorithm.

We tested this approach by simulating practitioner's feedback using the ISIC 2018 dataset. Our experiments show that our SENN is able, thanks to users feedback, to significantly increase its explanation power whithout compromising its classification accuracy, thus potentially increasing the trust of practitioners into computer-assisted diagnosis systems.



%
The main limitation of this approach is that it is applicable only when saliency maps offer a resolution that is high enough to provide users with a significantly fine-grained explanatory picture. Unfortunately, this is not the case in very deep neural networks, where the last convolutional stage (usually the preferred one to extract saliency maps) is composed by a high number of filters, but very few convoluted areas.
Hence, at the moment, the trade-off is to use less classification accurate networks with higher resolution explanation maps, in order to include humans in the correction freedback loop.

Future work will focus on testing this approach with the involvement or real practitioners and on different datasets, to verify how well these results generalize to other domains.

%
%
%
\bibliographystyle{splncs04}
\bibliography{ms}

\end{document}